\documentclass{bmvc2k}

\title{3D Structure-guided Network for Tooth Alignment in 2D Photograph}

\addauthor{Yulong Dou}{douyl2023@shanghaitech.edu.cn}{1}
\addauthor{Lanzhuju Mei}{meilzhj2023@shanghaitech.edu.cn}{1}
\addauthor{Dinggang Shen}{dgshen@shanghaitech.edu.cn}{1,2,3}
\addauthor{Zhiming Cui}{cuizhm@shanghaitech.edu.cn}{1}

\addinstitution{
 School of Biomedical Engineering\\
 ShanghaiTech University\\
 Shanghai 201210, China
}
\addinstitution{
 Shanghai United Imaging Intelligence Co., Ltd.\\
 Shanghai 200230, China 
}
\addinstitution{
 Shanghai Clinical Research and Trial Center\\
 Shanghai 201210, China
}

\runninghead{Yulong.Dou ET AL}{Tooth Alignment in photograph}

\usepackage{capt-of}
\usepackage{multirow}
\usepackage{amssymb}
\usepackage{diagbox}
\usepackage{mathrsfs}
\begin{document}

\maketitle

\begin{abstract}
Orthodontics focuses on rectifying misaligned teeth (i.e., malocclusions), affecting both masticatory function and aesthetics. However, orthodontic treatment often involves complex, lengthy procedures. As such, generating a 2D photograph depicting aligned teeth prior to orthodontic treatment is crucial for effective dentist-patient communication and, more importantly, for encouraging patients to accept orthodontic intervention. In this paper, we propose a 3D structure-guided tooth alignment network that takes 2D photographs as input (e.g., photos captured by smartphones) and aligns the teeth within the 2D image space to generate an orthodontic comparison photograph featuring aesthetically pleasing, aligned teeth. Notably, while the process operates within a 2D image space, our method employs 3D intra-oral scanning models collected in clinics to learn about orthodontic treatment, i.e., projecting the pre- and post-orthodontic 3D tooth structures onto 2D tooth contours, followed by a diffusion model to learn the mapping relationship. Ultimately, the aligned tooth contours are leveraged to guide the generation of a 2D photograph with aesthetically pleasing, aligned teeth and realistic textures. We evaluate our network on various facial photographs, demonstrating its exceptional performance and strong applicability within the orthodontic industry.
\end{abstract}

\section{Introduction}\label{sec:intro}
\begin{figure}[t]
  \centering
  \setlength{\abovecaptionskip}{-10pt}
  \setlength{\belowcaptionskip}{-15pt}
  \includegraphics[width=0.163\textwidth]{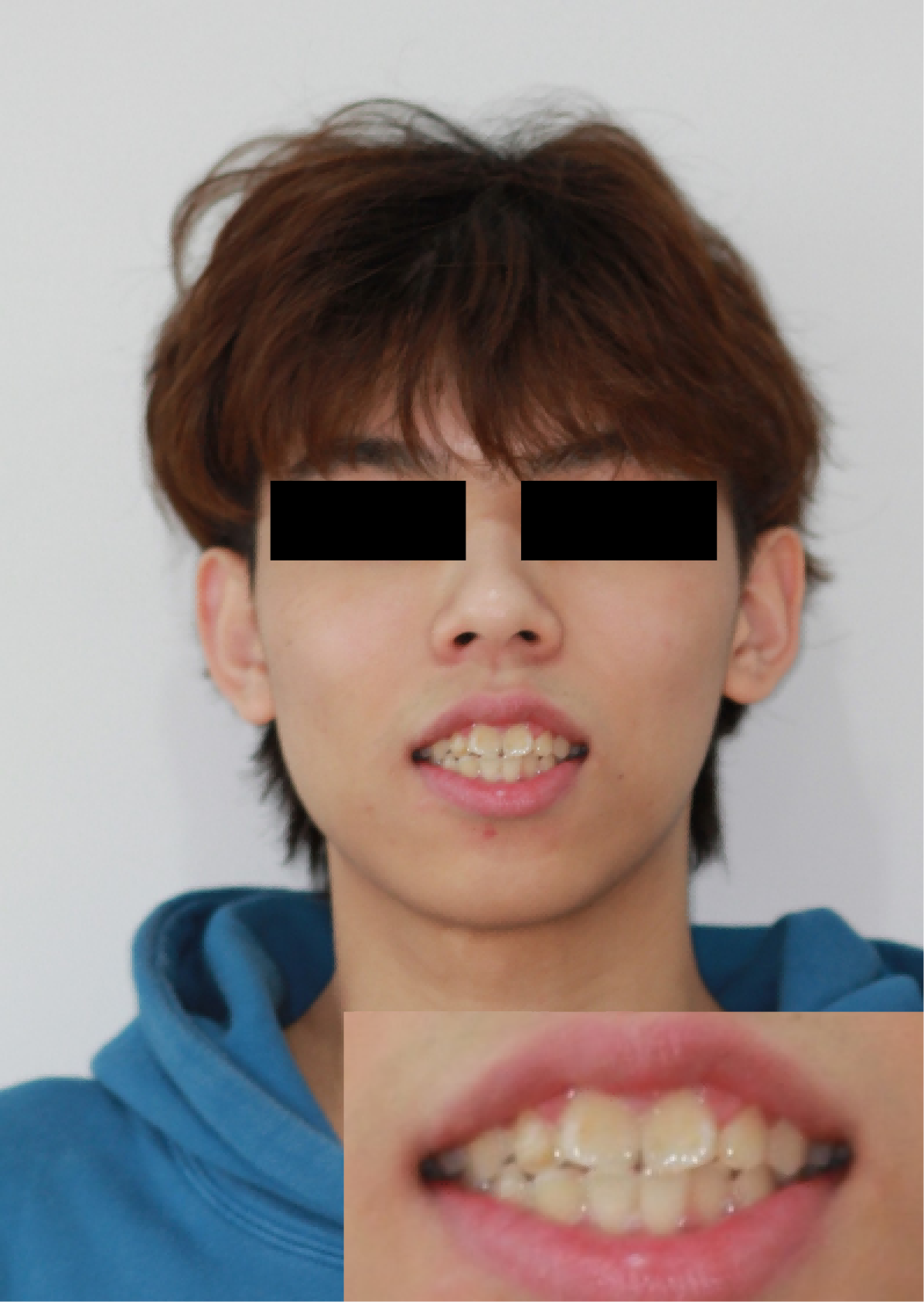}
  \includegraphics[width=0.163\textwidth]{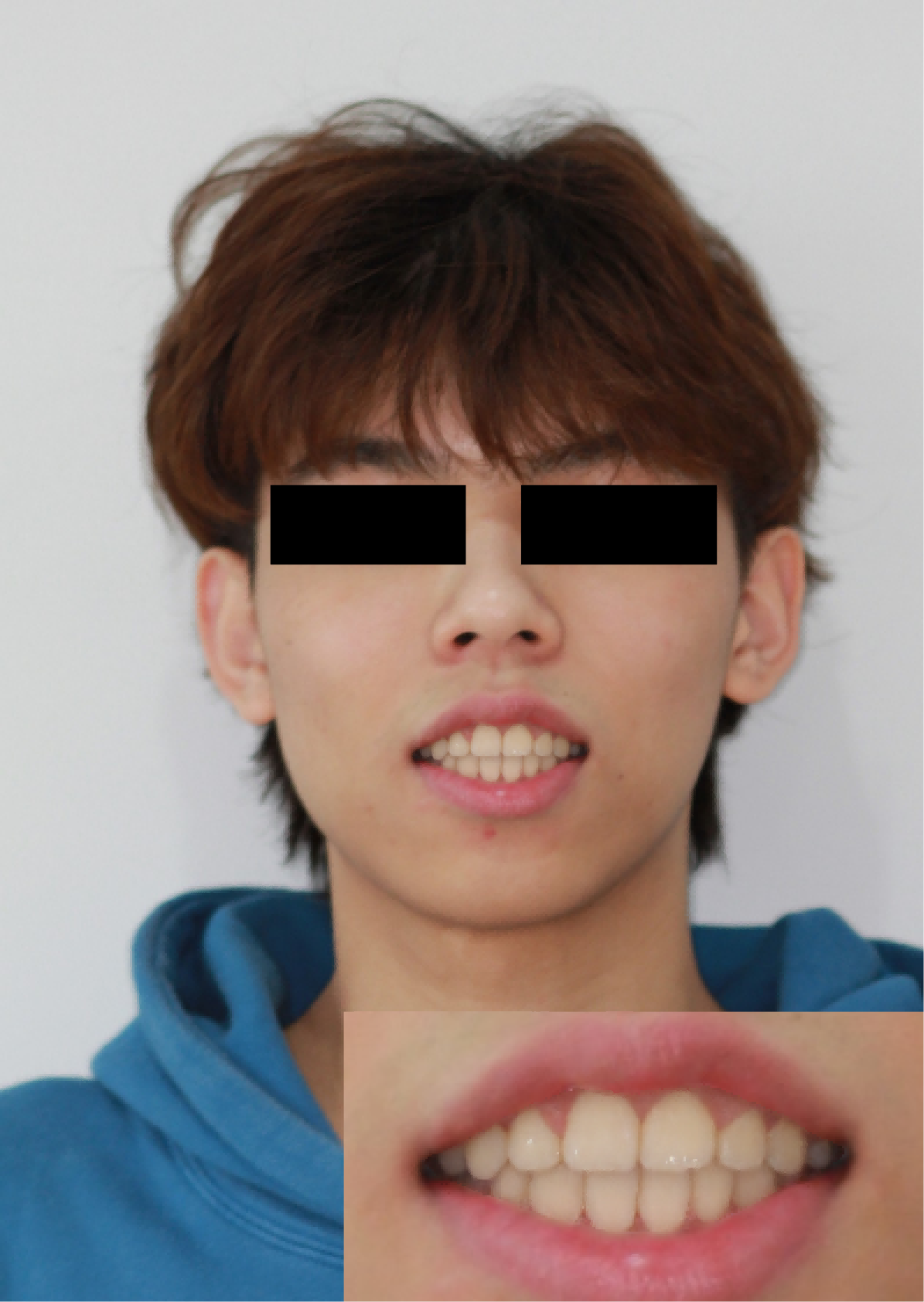}\hfill
  \includegraphics[width=0.163\textwidth]{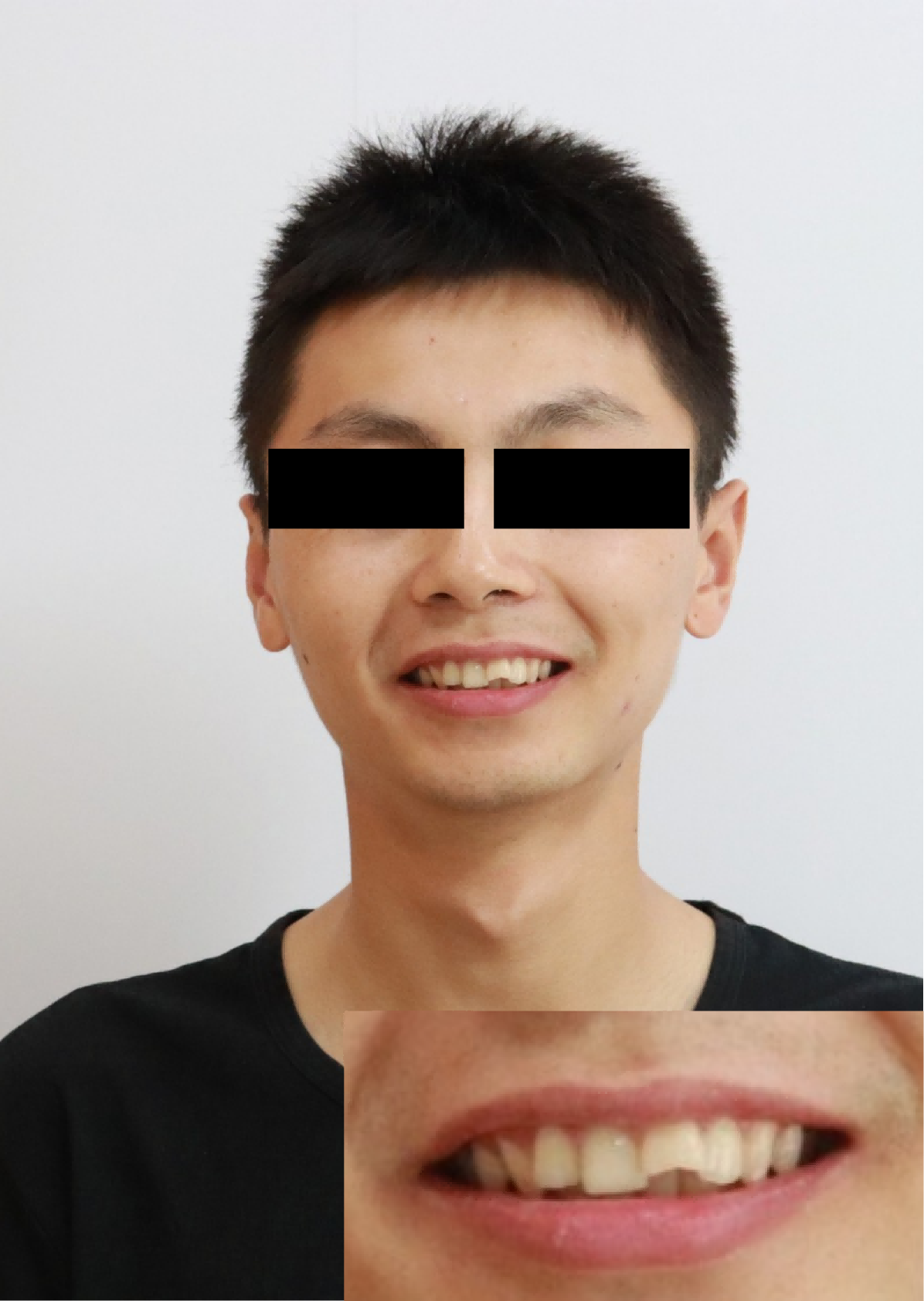}
  \includegraphics[width=0.163\textwidth]{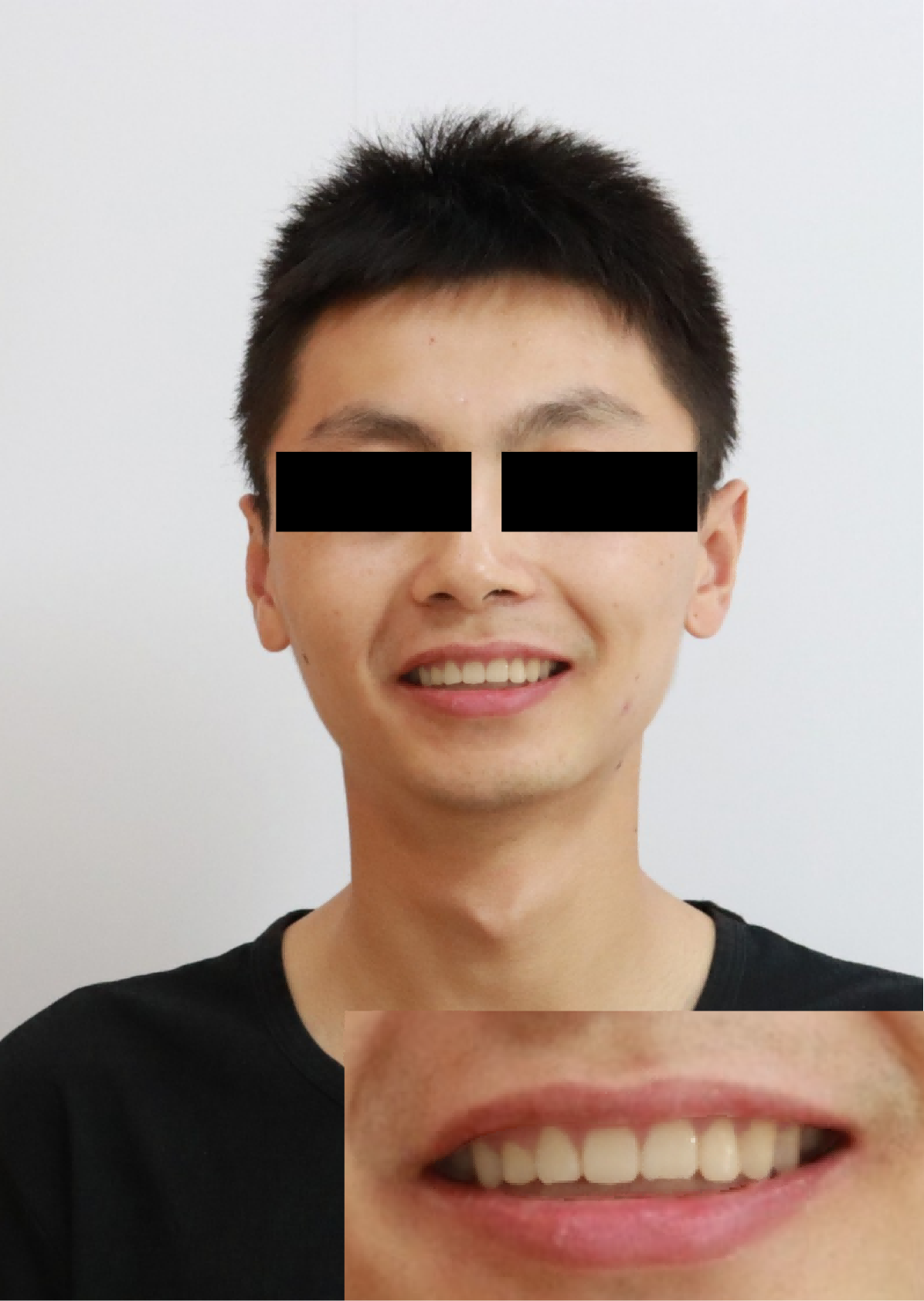}\hfill
  \includegraphics[width=0.163\textwidth]{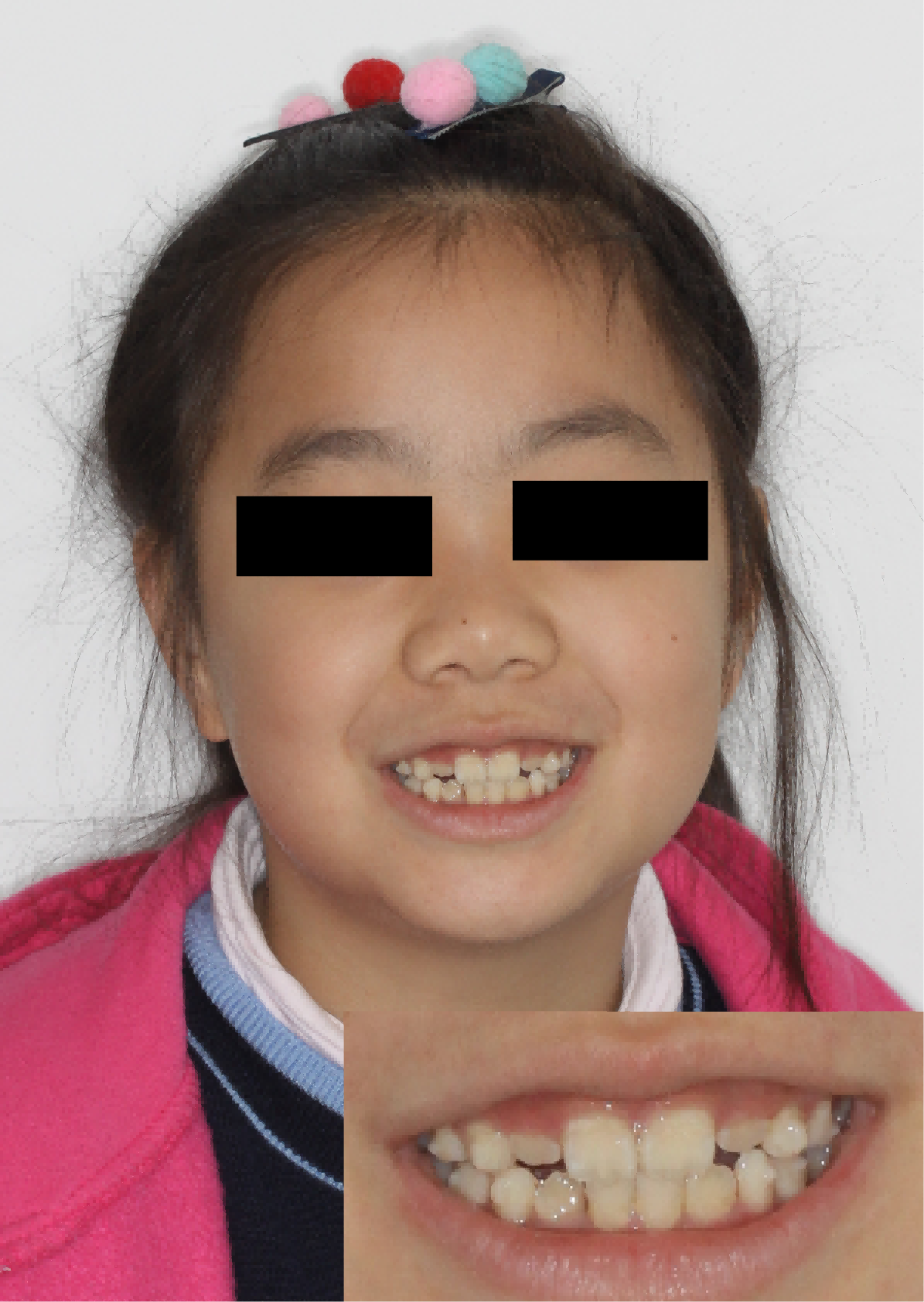}
  \includegraphics[width=0.163\textwidth]{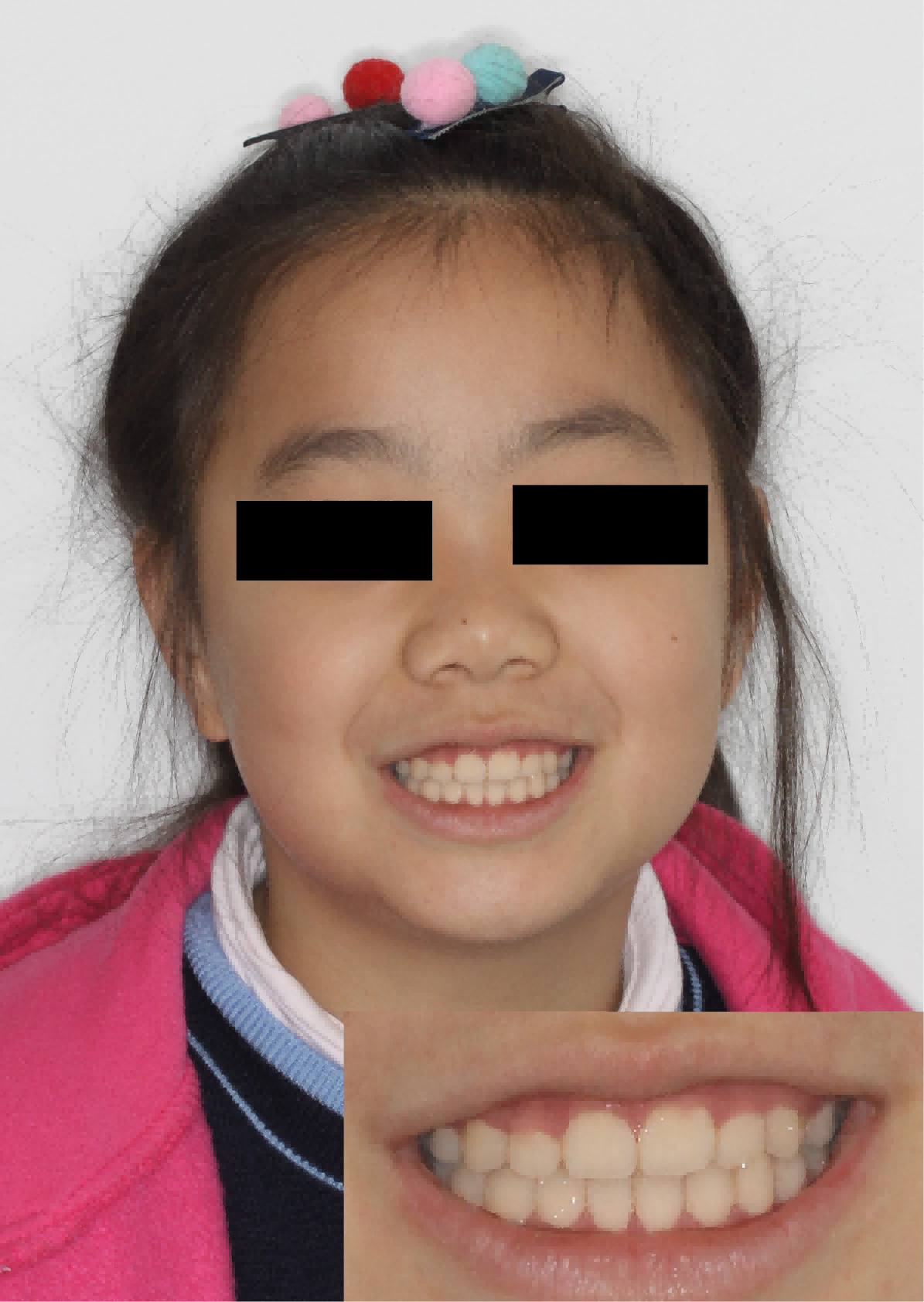}\hfill
  \includegraphics[width=0.163\textwidth]{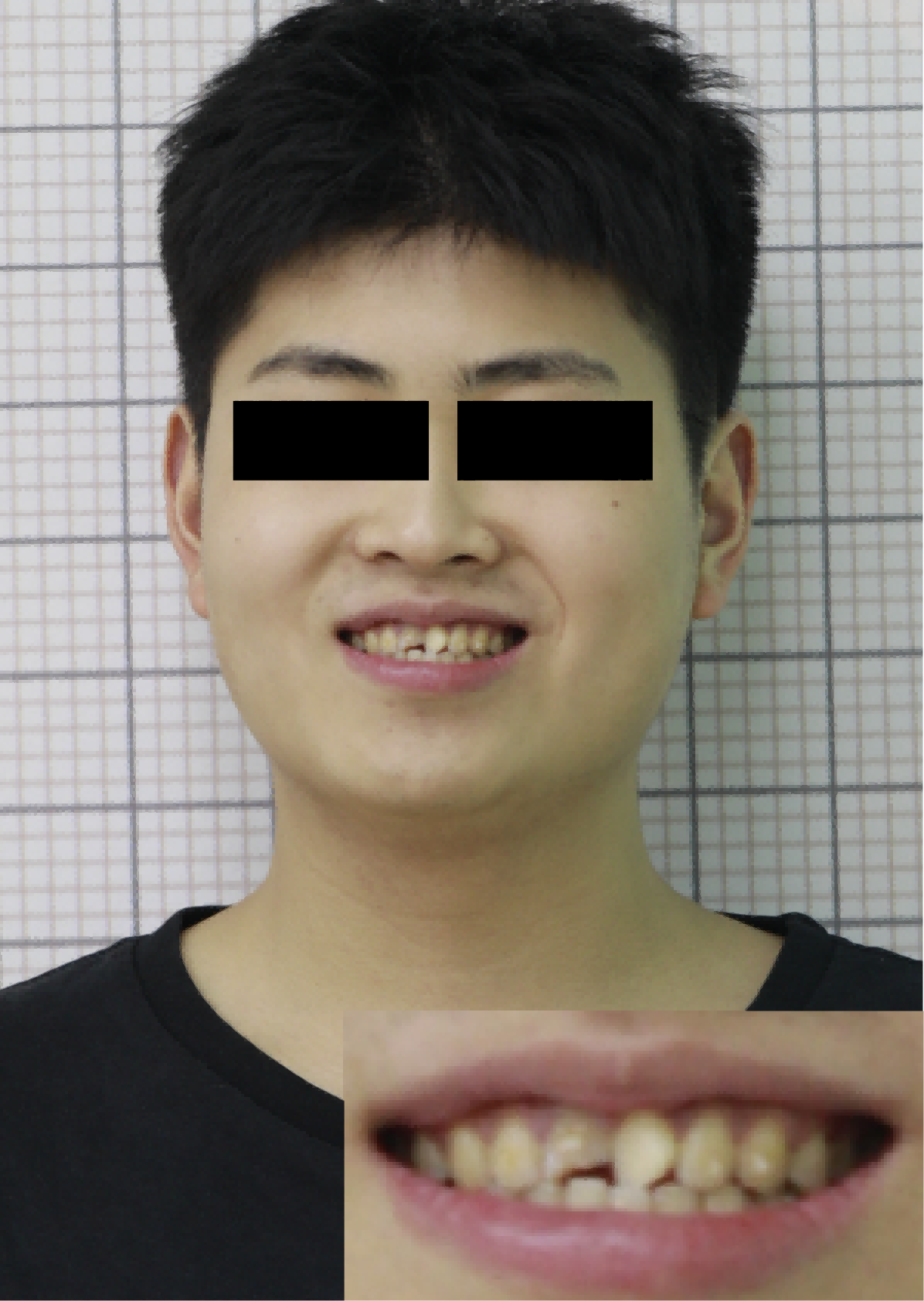}
  \includegraphics[width=0.163\textwidth]{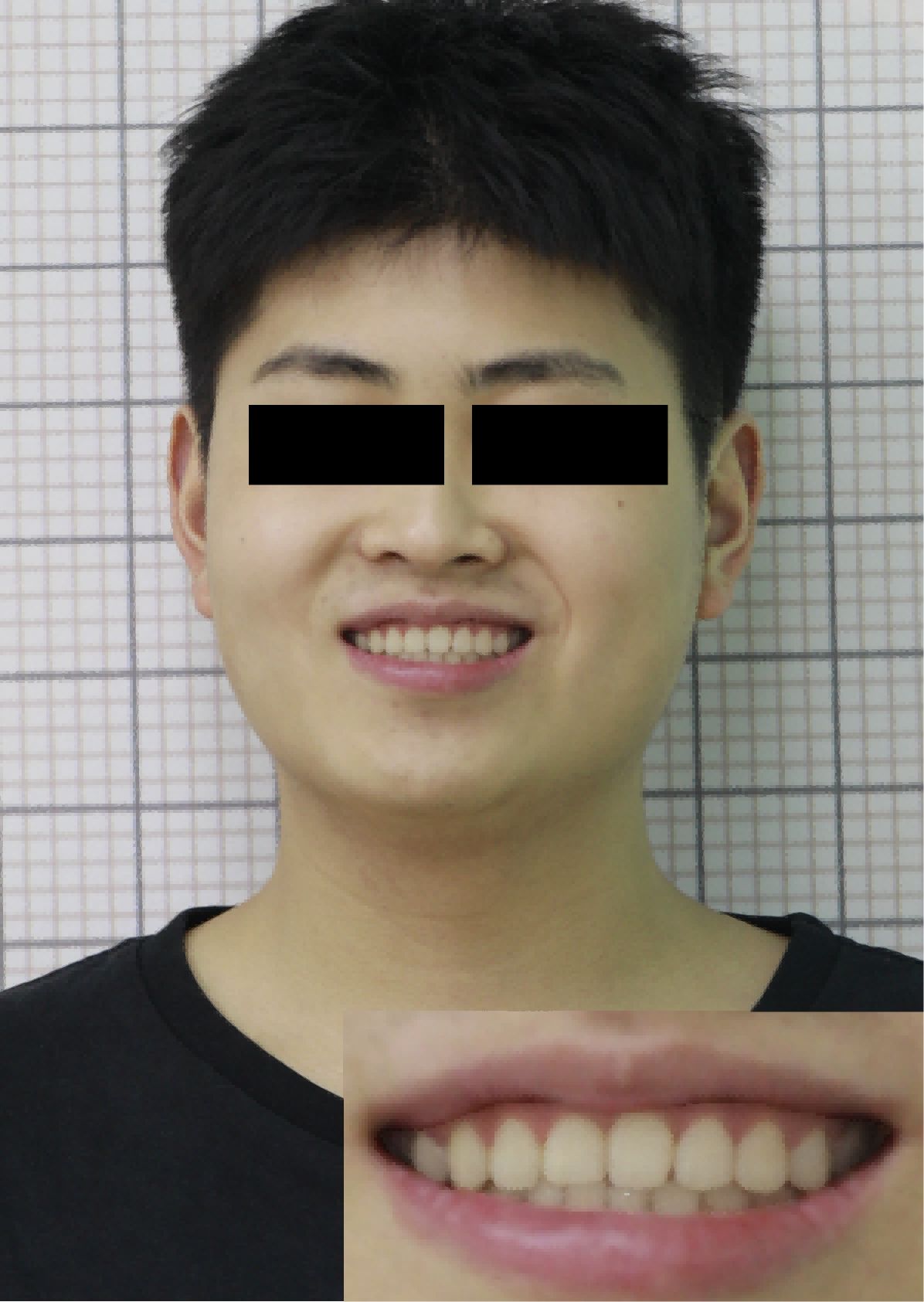}\hfill
  \includegraphics[width=0.163\textwidth]{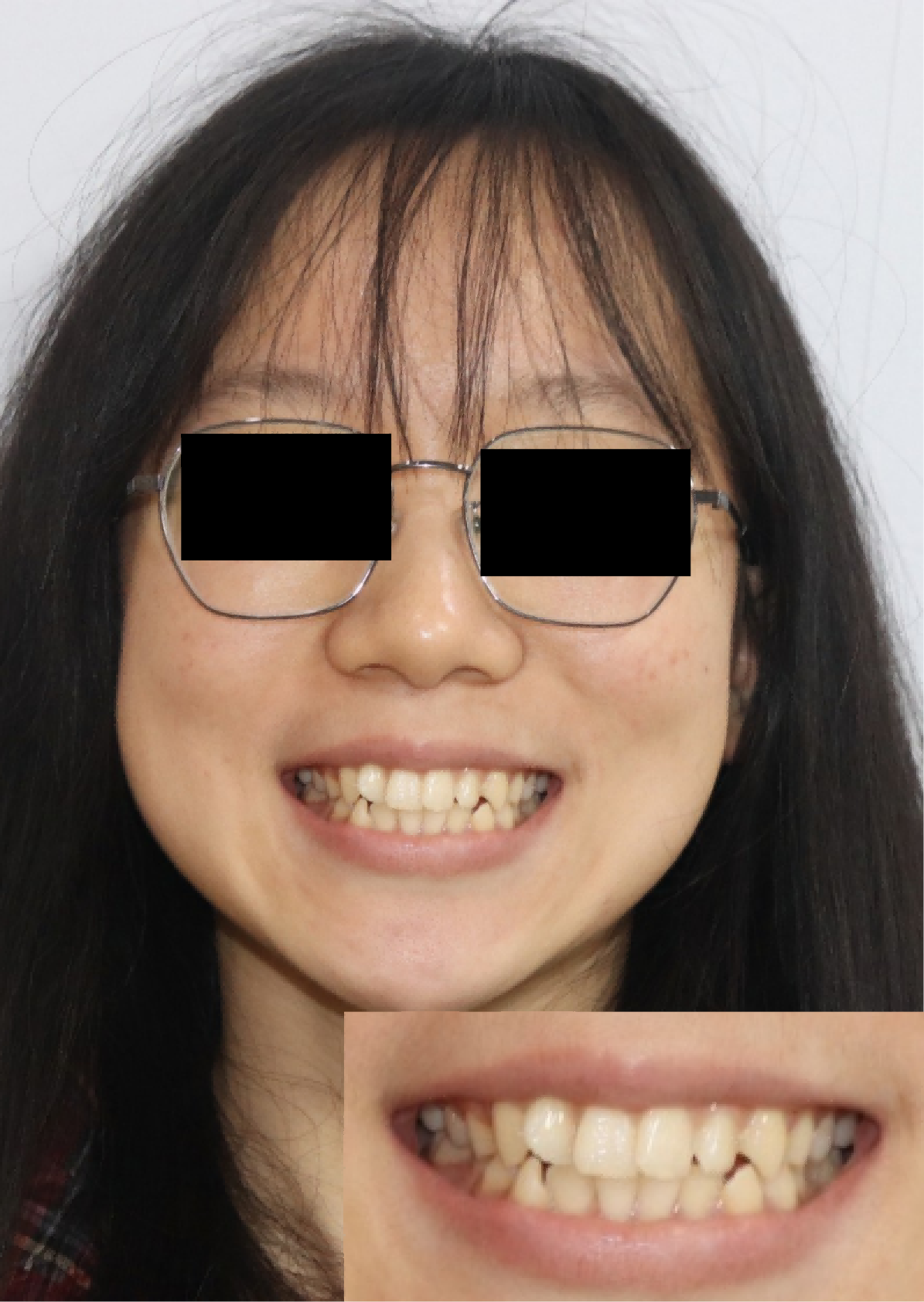}
  \includegraphics[width=0.163\textwidth]{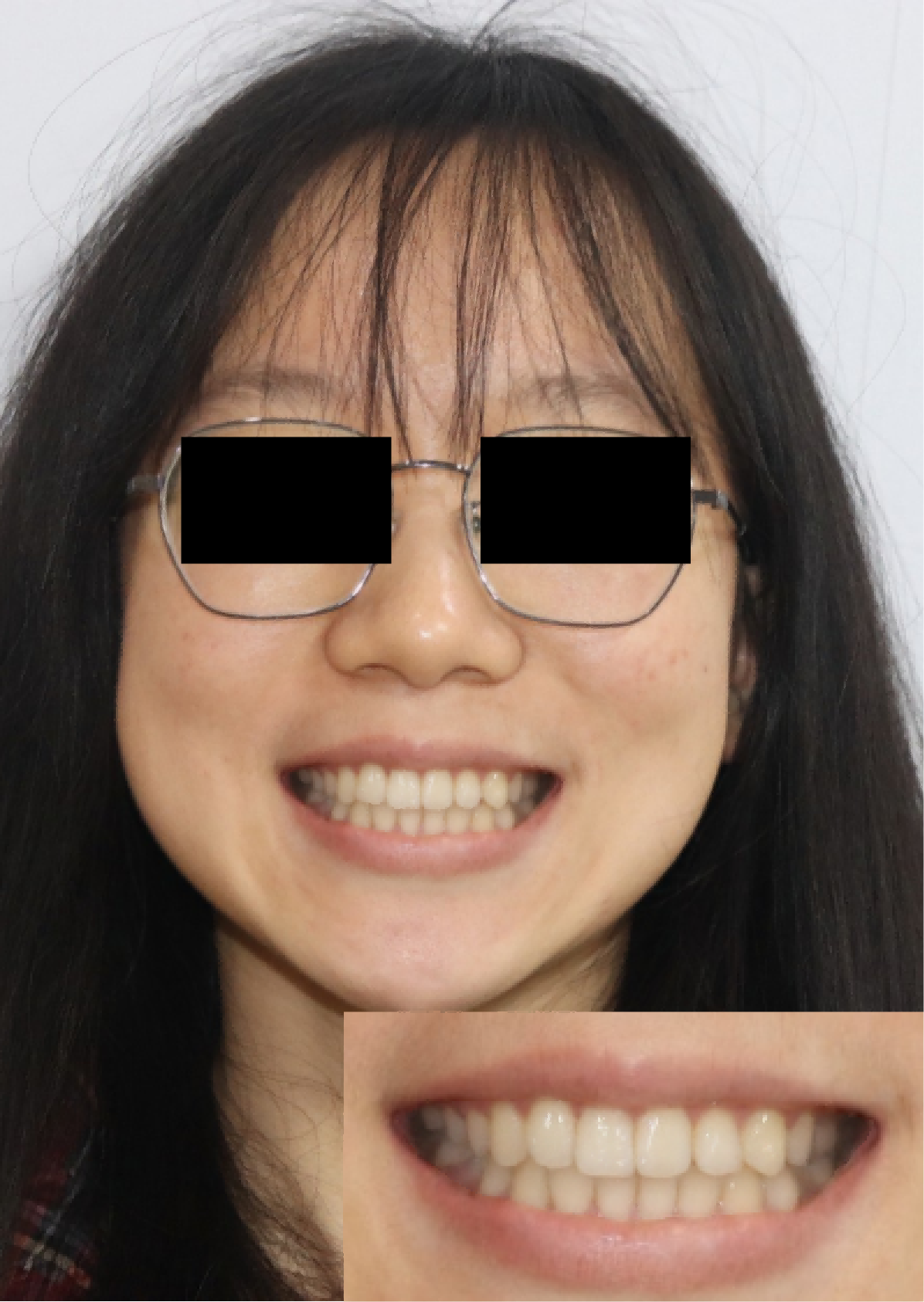}\hfill
  \includegraphics[width=0.163\textwidth]{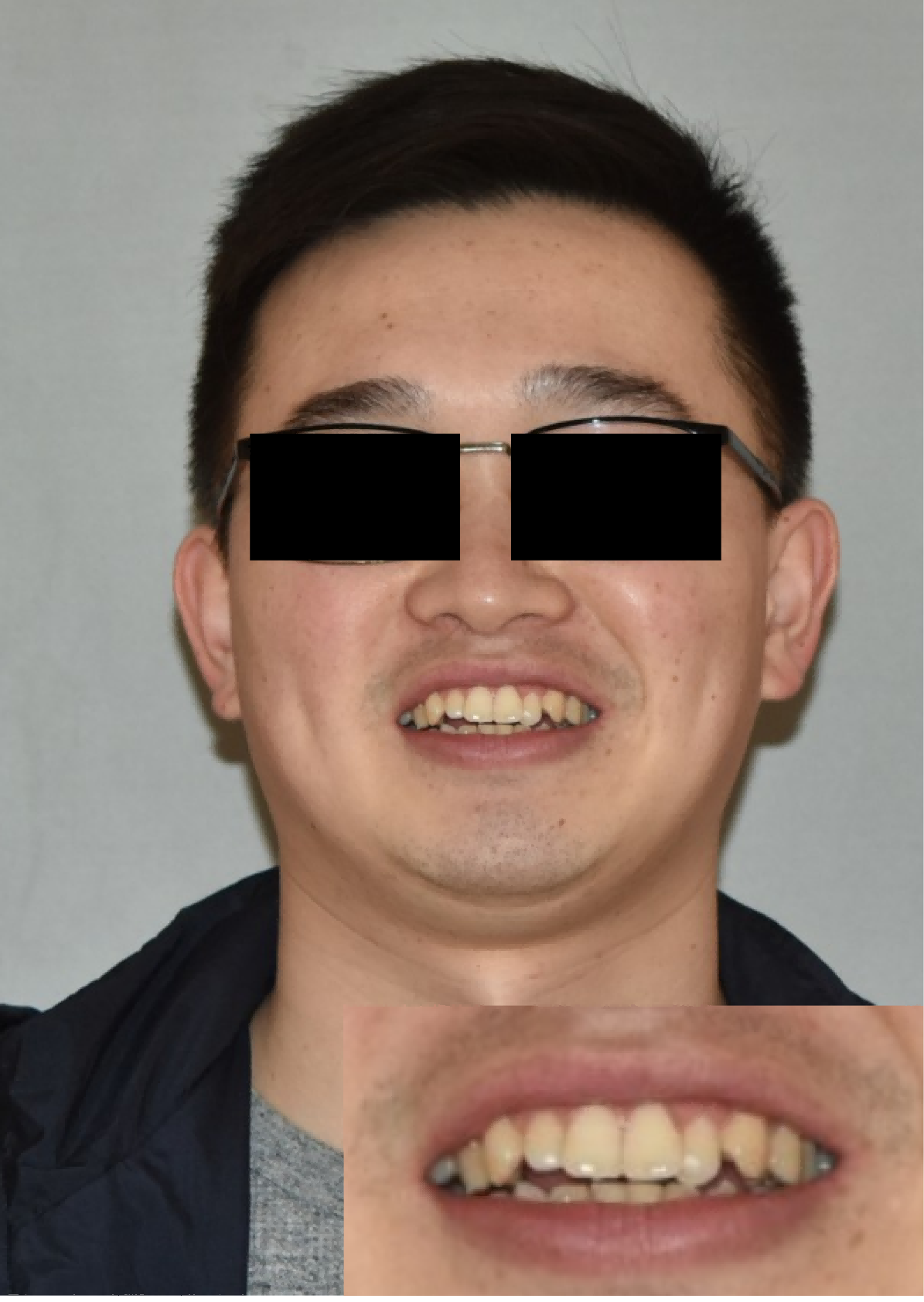}
  \includegraphics[width=0.163\textwidth]{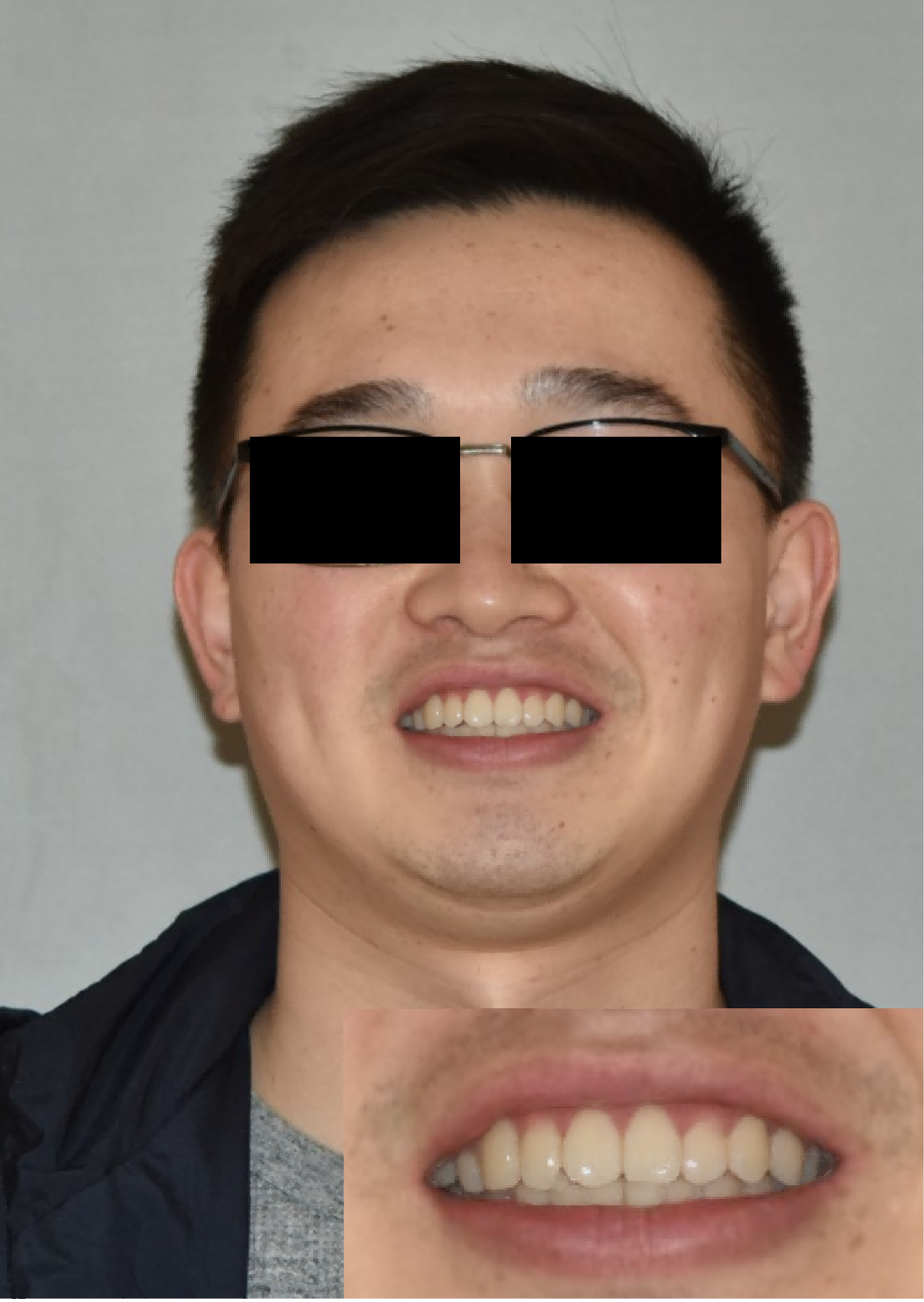}\hfill
  \caption{Orthodontic comparison photographs. For each case, we show the facial photograph with misaligned teeth (left) and the facial photograph with well-aligned teeth generated by our network (right), and the image in the lower right corner is a zoom-in of mouth region.}
  \label{fig:result_vis}
\end{figure}

Orthodontic treatment is an effective remedy for correcting tooth misalignment (i.e., malocclusions). It is estimated that over $90\%$ people suffer from malocclusion problems with various degrees\citep{bernhardt2019new}, and most of people can benefit from orthodontic intervention. This treatment not only helps prevent oral diseases at a physiological level, but also significantly boosts patients' confidence, enhancing their psychological well-being\citep{peng2017shen}. However, the complexity of orthodontic procedure, which often spans several months or even years, can deter individuals from seeking treatment. Hence, the generation and visualization of potential post-treatment facial photographs with aesthetical teeth becomes crucial. Such predictive imaging not only engages and motivates patients but also fosters more effective communication between orthodontists and their patients.

In clinical practice, visualizing patients’ appearance after orthodontic treatment is referred to "Visual Treatment Objective" (VTO). This is typically performed on X-Ray images by deforming soft tissues and skeleton based on detected landmarks\citep{mclaughlin1999dental,power2005dolphin}. However, this operation leaves the teeth's appearance unaltered,, making it challenging for patients to make a realistic comparison. In this study, our objective is to take 2D photograph as an input (e.g., photos captured by smartphones), and directly generate the "Orthodontic Comparison Photograph" with aligned teeth and realistic textures, as shown in Figure \ref{fig:result_vis}. Note that the generated photograph should follow the unique tooth alignment property of each patient in real-world treatment, instead of simplistic Photoshop approach with template teeth\citep{sundar2018ten}. 

Currently, significant advancements in deep learning, particularly in generative networks, have achieved promising results in computer vision community. However, most of these models heavily rely on paired images, that is not suitable for our task.  This is primarily because collecting paired pre- and post-orthodontic facial photographs is challenging due to the long-term orthodontic procedure and changes in facial appearance over time. Furthermore, the 2D photograph does not provide the 3D structure of teeth. Thus, how to learn the clinical knowledge of tooth alignment, defined on 3D tooth models, from 2D photographs is also a significant challenge. 

In this paper, we propose a 3D structure-guided network for tooth alignment in 2D photograph. The key idea is to learn the clinical tooth alignment knowledge defined on the 3D intra-oral scanning models\citep{kihara2020accuracy}, and apply the learned property to guide the 2D post-orthodontic photograph generation. Specifically, we begin by collecting a set of paired pre- and post-orthodontic intra-oral scanning tooth models in clinics, and render\citep{pharr2016physically} them onto the oral area of a 2D facial photograph. In this way, we can obtain the paired pre- and post-orthodontic tooth contours in 2D photograph (as shown in Figure \ref{fig:pipeline}). Then, a Diffusion Model\citep{ho2020denoising} is applied to learn tooth alignment knowledge, i.e., generating post-orthodontic tooth contours with the input of pre-orthodontic tooth contours, derived from 3D tooth models. Note that only the tooth structures are captured, without any texture information. In the inference process, we can directly take the tooth contours segmented from 2D facial photograph. Finally, guided by the aligned tooth contours, we employ another Diffusion Model to generate a realistic 2D photograph with aligned teeth. In particular, To enhance similarity with the patient's original appearance, we incorporate facial skin color and intra-oral highlights into the generation process, accounting for texture and lighting information. In the experiment, we collect a large number of photographs from patients suffering from malocclusion problems with various degrees, and achieve superior performance compared to the state-of-the-art methods, including GAN\citep{goodfellow2020generative} and Diffusion Models. Furthermore, we also conduct a user study to validate the alignment and authenticity performance of our algorithm, demonstrating its potential applicability within orthodontic industry.

\vspace{-0.2em}
\section{Related Work}\label{sec:related}
{\bf Digital Orthodontics.} Digital orthodontics employs digital imaging technologies such as intra-oral scanning\citep{kihara2020accuracy}, CBCT\citep{de2009cone}, and panoramic radiograph\citep{angelopoulos2004digital} to provide dentists with information about the structure and occlusion of patients' teeth. This helps dentists with pre-treatment diagnosis and orthodontic treatment planning. A variety of emerging techniques have been introduced in related fields, including tooth segmentation\citep{cui2019toothnet,cui2022fully}, 3D tooth reconstruction\citep{wu2016model,wirtz2021automatic}, and 3D tooth arrangement\citep{wei2020tanet}. In terms of orthodontic comparison photographs, \citet{lingchen2020iorthopredictor} have introduced iOrthoPredictor which can synthesize an image of well-aligned teeth based on a patient's facial photograph and an additional input of the patient's 3D dental model. \citet{chen2022orthoaligner} have introduced OrthoAligner which needs only a facial photograph but no 3D dental model as input, by introducing the concept of StyleGAN inversion. But OrthoAligner is limited in that it only uses facial photographs to learn tooth transformation, without utilizing information from 3D dental models.

\vspace{0.2em}
\noindent {\bf Image Generation.} Image generation is a field of research in computer vision that aims to generate new digital images by using algorithms or models from scratch or by modifying existing images. Several models have been proposed for image generation, including GAN\citep{goodfellow2020generative}, VAE\citep{kingma2013auto}, Diffusion Model\citep{ho2020denoising}. Specifically, GAN simultaneously trains the generator and discriminator to generate more realistic images. Many models based on GAN have been proposed, such as unsupervised StyleGAN\citep{karras2019style}, and supervised Pix2pix GAN\citep{isola2017image}. VAE is a generative model that uses variational inference for sampling from probability distributions. \citet{ho2020denoising} propose Diffusion Model based on Score Matching\citep{hyvarinen2005estimation} and Denoising Autoencoder\citep{vincent2011connection}, and elaborate on its mathematical principles. Diffusion Model is a generative model that utilizes a forward process of step-by-step adding noise and a backward denoising process to generate high-quality images. \citet{choi2021ilvr} propose a reference-guided conditional Diffusion Model that fine-tunes the backward denoising process. \citet{singh2022conditioning} introduce condition noise to navigate Diffusion Model. \citet{saharia2022palette} propose an image-to-image Diffusion Model guided by condition image.

\section{Method}\label{sec:method}
\subsection{Overview}
Overall, our goal is to design a tooth alignment network that incorporates 3D structural information derived from intra-oral scanning models, which is essential for clinical orthodontic treatment, and guide the orthodontic comparison photograph generation. 

We have pre-orthodontic intra-oral scanning models $S=\{S_1,S_2,...,S_N\}$, post-orthodontic intra-oral scanning models $\hat{S}=\{\hat{S_1},\hat{S_2},...,\hat{S_N}\}$ of same patients collected in clinics, and unpaired facial photographs $I=\{I_1,I_2,...,I_M\}$ collected by smartphones.
Given that the facial photographs $I$ and 3D intra-oral scanning models $S,\hat{S}$ in our dataset are not paired,  we design a module, named $Align\text{-}Mod$, for tooth alignment that can still incorporate 3D structural information from intra-oral scanning models as guidance. 
This module randomly selects the pre- and post-orthodontic intra-oral scanning models (i.e., $S_r\in S$ and $\hat{S_r}\in \hat{S}$) for an unpaired facial photograph (i.e., $I_r\in I$), and makes a coarse 2D-3D registration between $S_r,\hat{S_r}$ and $I_r$, respectively.
Then, the 3D tooth structures are projected onto the 2D facial photograph to obtain pre-orthodontic tooth contours $C_r\in \mathbb{R}^{128\times 256}$ and post-orthodontic tooth contours $\hat{C_r}\in \mathbb{R}^{128\times 256}$. 
{\bf In this way, our $Align\text{-}Mod$ module can learn the tooth transformation $\mathcal{T}(\cdot)$, which represents the clinical orthodontic knowledge derived from the 3D intra-oral scanning models.} 

In addition to the pre-trained tooth alignment module, we also design a {\bf segmentation module, named $Segm\text{-}Mod$, to locate the mouth region and segment tooth contours $C$ from facial photographs $I$}, and a {\bf generation module, named $Gen\text{-}Mod$, to generate the facial image with aesthetically pleasing teeth.}

\begin{figure}[htbp]
    \setlength{\abovecaptionskip}{-10pt}
    \setlength{\belowcaptionskip}{-10pt}
    \centering
    \includegraphics[width=1\textwidth]{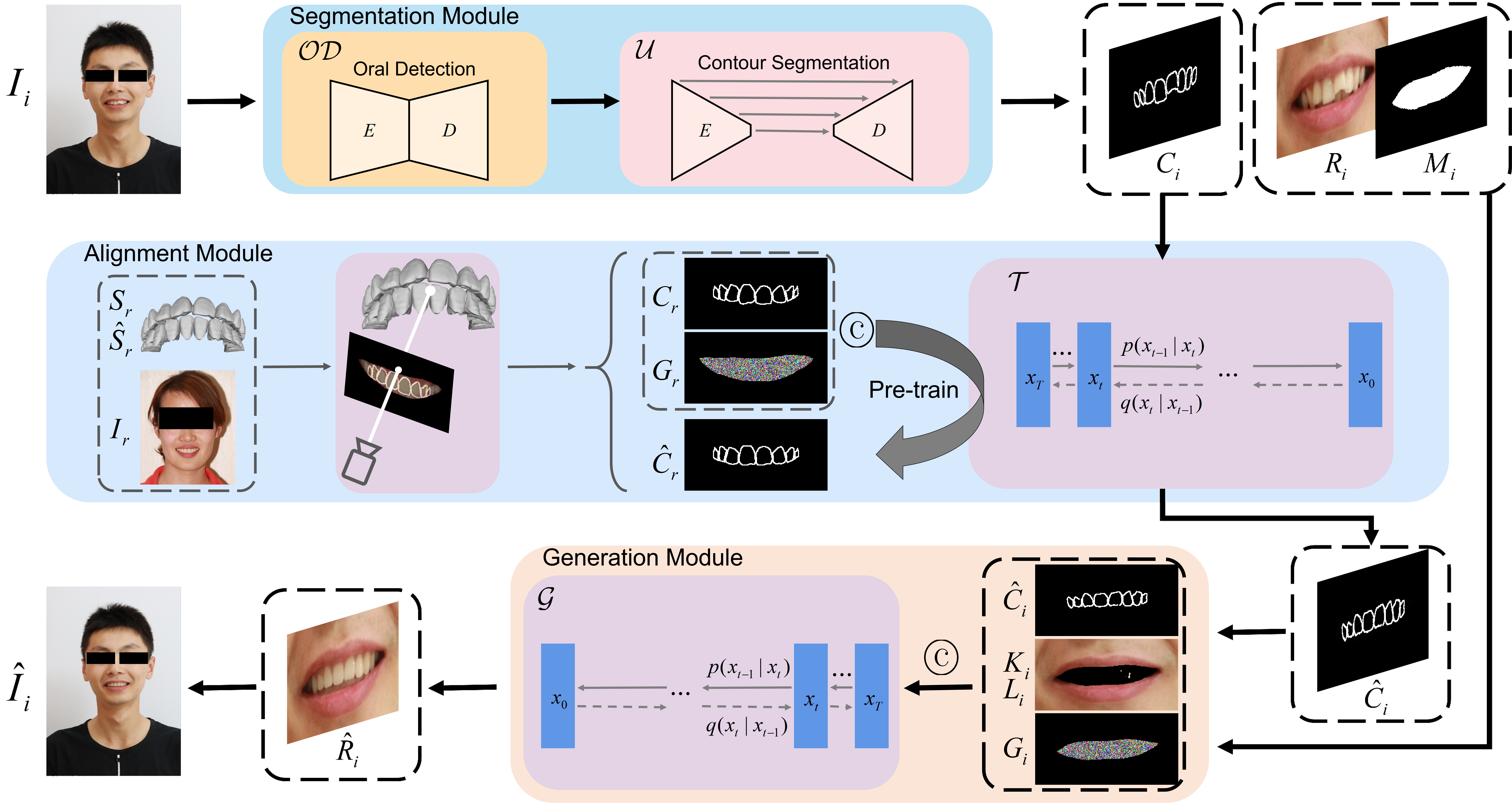}
    \caption{Overall pipeline. When a facial photograph is input into our network, it first goes through $Segm\text{-}Mod$ to obtain oral mask, mouth region and tooth contours. Then it enters Pre-trained $Align\text{-}Mod$ to predict well-aligned tooth contours, and finally goes through $Gen\text{-}Mod$ to generate a facial photograph with well-aligned teeth.}
    \label{fig:pipeline}
\end{figure}

In summary, the three modules designed in this framework are shown in Figure \ref{fig:pipeline}.

\subsection{Segmentation Module}
To begin with, $Segm\text{-}Mod$ needs to detect the position of face\citep{kazemi2014one,schroff2015facenet} and obtain a standardized face $F_i \in \mathbb{R}^{512\times 512}$ from any given facial photograph $I_i\in I \subseteq \mathbb{R}$. As shown in Figure \ref{fig:pipeline}, to accurately locate the mouth, we propose an oral detection network $\mathcal{OD}(\cdot)$\citep{yu2018bisenet,lee2020maskgan} to segment the oral mask $M_i\in \mathbb{R}^{128\times 256}$ and crop the mouth region $R_i\in \mathbb{R}^{128\times 256}$ from the standardized face $F_i$. 
Then, to obtain tooth contours $C_i\in \mathbb{R}^{128\times 256}$ which contains structural information of teeth, we employ a commonly-used supervised segmentation network, U-Net\citep{ronneberger2015u}, for segmenting tooth contours $C_i$ from the mouth region $R_i$. The formulation of process of $Segm\text{-}Mod$ is as:
\begin{equation} \label{eq:segm}
\begin{aligned}
  R_i,\ M_i &= \mathcal{OD}(I_i), \\
  C_i = \mathcal{U}(R_i,\ M_i&),\ \ \forall i=1,2,...,N,
\end{aligned}
\end{equation}
where $\mathcal{OD}(\cdot)$ is the oral detection network performed in 2D facial photograph $I_i$, $\mathcal{U}(\cdot)$ denotes the U-Net-based contour segmentation network, and $C_i,\ R_i,\ M_i$ are the pre-orthodontic tooth contours, mouth region and oral mask obtained by $Segm\text{-}Mod$, respectively.

To train the network, we employ the Dice Loss\citep{milletari2016v} and Weighted Cross-Entropy Loss\citep{rubinstein1999cross}. Given the imbalanced area between foreground (tooth contours) to background, Dice Loss has excellent performance in situations with severe imbalance and focuses on learning the foreground area. Furthermore, Weighted Cross-Entropy Loss can address the imbalance problem by adjusting the weighted proportion, thus making it a suitable complement to Dice Loss. Our designed loss function is defined as:
\begin{equation} \label{eq:loss}
  \mathcal{L}  = w_{dice} \cdot \mathcal{L}_{dice} + w_{ce} \cdot \mathcal{L}_{ce}.
\end{equation}

\subsection{Alignment Module}
One of the most innovative aspects of our method is that we can incorporate structural information from 3D intra-oral scanning models $S,\hat{S}$ into $Align\text{-}Mod$, which are essential for clinical orthodontic treatment. We employ a 3D-to-2D Render to project the 3D intra-oral scanning models $S,\hat{S}$ into the oral area of the 2D facial photographs, as opposed to the approach used by \citet{zheng2020key,wirtz2021automatic}  to reconstruct 3D dental models from multi-view tooth photographs. Furthermore, we design a conditional Diffusion Model-based network for learning the clinical orthodontic knowledge $\mathcal{T}(\cdot)$ in the space of tooth contours $C_r,\hat{C_r}$ obtained by Render.

\subsubsection{Render} \label{subsubsec:render}
Since the 3D intra-oral scanning models $S,\hat{S}$ and facial photographs $I$ are collected from different environments and sources, i.e., one is in clinics and another is from smartphones in daily life, we cannot perform precise 3D-2D registration through rigid transformation. Fortunately, precise registration is not necessary for our task as our purpose is to create paired tooth contours from the intra-oral scanning models. Therefore, to obtain tooth contours $C_r,\hat{C_r} \in \mathbb{R}^{128\times 256}$, we perform coarse registration based on the landmarks between the coordinates of tooth cusp points of central and lateral incisors in both render-used intra-oral scanning models $S_r\in S,\hat{S_r}\in \hat{S}$ and facial photograph $I_r\in I$. 

We use Numerical Optimization\citep{tapley1967comparison} to perform coarse registration and then obtain 2D tooth contours by projection. The essential principle of projecting 3D to 2D is to solve the camera parameters. As shown in Equation \ref{eq:camera}, 
\begin{equation} \label{eq:camera}
  \rho m=(K R^{T}|-K R^{T}C)\left(\begin{array}{c}{{M}}\\ {{1}}\end{array}\right),
\end{equation}
where $M$ represents the coordinate of the point in world coordinate system, denoted as $M = (X,Y,Z)^T$, and $m$ represents the coordinate of the corresponding point in pixel coordinate system, denoted as $m = (u,v,1)^T$. $\rho$ is the projection depth, $C$ is the position of camera, $R$ is the rotation matrix that represents camera pose, and $K$ is the matrix of intrinsic parameters\citep{hartley2003multiple}. 
Specifically, we use the four paired tooth cusp points mentioned above on both the facial photo and the intra-oral scanning model to derive their coordinates, represented as $m$ and $M$. Subsequently, we can resolve camera parameters, primarily the unknown variables in $K$ and $C$, given the other known parameters.

\subsubsection{Tooth Transformation}
Once we render 3D intra-oral scanning models $S_r,\hat{S_r}$ onto the 2D facial photograph $I_r$ and obtain pre- and post-orthodontic tooth contours  $C_r,\hat{C_r}$ as mentioned above, we can then learn the $\mathcal{T}(\cdot)$, i.e., clinical orthodontic knowledge. We employ a network based on image-to-image Diffusion Model\citep{saharia2022palette}, as shown in Figure \ref{fig:pipeline}. To emphasize the generation of oral regions, we introduce Gaussian noise $G_r\in \mathbb{R}^{128\times 256}$ generated within the oral mask $M_r$. We concatenate the pre-orthodontic tooth contours $C_r$ with the Gaussian noise $G_r$ to form the condition information, which serves as guidance for our diffusion model. Therefore, the formula for $\mathcal{T}(\cdot)$ during the pre-trained process is given as:
\begin{equation}  \label{eq:trans}
  \hat{C_r} = \mathcal{T}(C_r \, \textcircled{c} \, G_r),
\end{equation}
where $\textcircled{c}$ denotes channel-wise concatenation.

Since we have pre-learned the $\mathcal{T}(\cdot)$, which represents the clinical orthodontic knowledge of tooth transformation, we can apply the learned knowledge to process the 2D tooth contours $C_i$ derived from $Segm\text{-}Mod$. Hence, we concatenate the tooth contours $C_i$, together with intra-oral Gaussian noise $G_i$, and then feed them into our diffusion model, expecting a reasonable prediction for well-aligned tooth contours $\hat{C_i}$. The inference process is formulated as: 
\begin{equation} \label{eq:align}
  \hat{C_i} = \mathcal{T}(C_i \, \textcircled{c} \, G_i),\ \ \forall i=1,2,...,N.
\end{equation}

\subsection{Generation Module} \label{subsec:generation}
After obtaining well-aligned tooth contours $\hat{C_i}$ through the tooth transformation $\mathcal{T}(\cdot)$  of our $Align\text{-}Mod$, we aim to generate a mouth region with realistic teeth $\hat{R_i}$ guided by $\hat{C_i}$. To achieve this, we adapt a conditional Diffusion Model-based generative network $\mathcal{G}(\cdot)$ in our $Gen\text{-}Mod$. We still introduce Gaussian noise $G_i\in \mathbb{R}^{128\times 256}$ generated within the oral mask $M_i$ to emphasize the generation of oral regions. Besides well-aligned tooth contours $\hat{C_i}$ and intra-oral Gaussian noise $G_i$ mentioned above, we additionally introduce intra-oral highlights $L_i\in \mathbb{R}^{128\times 256}$ and facial skin color $K_i\in \mathbb{R}^{128\times 256}$, which are helpful for generating more realistic tooth color and environmental lighting. Then, four of them are concatenated together as the condition information to guide our generation model.

In terms of intra-oral highlights, we employ Contrast Limited Adaptive Histogram Equalization (CLAHE) \citep{zuiderveld1994contrast} and Thresholding, aimed to enhance image contrast and detect highlights within the oral region. Specifically, we utilize CLAHE as in Equation \ref{eq:CLAHE} to improve local contrast in mouth region $R_i$ by using a histogram equalization approach with a specified contrast limit of 5 in $20\times20$ local window, thus preventing over-amplification of noise in flat areas while enhancing contrast in textured areas, defined as:
\begin{align} \label{eq:CLAHE}
g_{xy} &= \frac{L-1}{S_{xy}} \sum_{z=0}^{f_{xy}} \frac{h(x,z)}{S_{xy}},
& f'_{xy} &= g_{xy} \times (L-1),
& \text{CLAHE}_{xy} &=
\begin{cases}
f'_{xy} & \text{if } f'_{xy} < L-1 \\
L-1 & \text{if } f'_{xy} \geq L-1 
\end{cases},
\\[-10pt] \notag
\end{align}
where $S_{xy}$ denotes size of the local window, $f_{xy}$ denotes pixel intensity at pixel $(x,y)$, $h(x,z)$ is histogram of pixel intensities in local window, $L$ means number of intensity level, $g_{xy}$ is gain factor, and $f'_{xy}$ is transformed pixel intensity.

Once we generate a mouth region with realistic teeth $\hat{R_i}$, we use the face and mouth position stored in $Segm\text{-}Mod$ to replace the oral region in the initial facial photograph, thereby obtaining a facial photograph with well-aligned and aesthetically pleasing teeth $\hat{I_i}$ for VTO (see Figure \ref{fig:pipeline}). The formulation of process of $Gen\text{-}Mod$ is shown as:
\begin{equation} \label{eq:gen}
  \hat{I_i} = \mathcal{G}(\hat{C_i}\, \textcircled{c}\, G_i\, \textcircled{c}\, L_i\, \textcircled{c}\, K_i),\ \ \forall i=1,2,...,N,
\end{equation}
where $\textcircled{c}$ denotes channel-wise concatenation of segmented tooth contours $\hat{C_i}$, Gaussian noise $G_i$, intra-oral highlights $L_i$ and facial skin color $K_i$. $\hat{I_i}$ is the predicted facial photograph with well-aligned and aesthetically pleasing teeth through the Diffusion Model-based generative network $\mathcal{G}(\cdot)$ mentioned in $Gen\text{-}Mod$.

\section{Experiments} \label{sec:experiments}
\subsection{Experiments Settings}
{\bf Dataset.}  Our dataset comprises 1367 facial photographs $I$, of which 1129 are used to train $Segm\text{-}Mod$ and $Gen\text{-}Mod$, 138 are used to create datasets through Render in $Align\text{-}Mod$, and the remaining 100 are reserved for testing our overall pipeline. For the 138 render-used facial photographs, we manually annotate the coordinates of tooth cusp points of central and lateral incisors in the upper jaw. 
Moreover, We have 1257 3D intra-oral scanning models $S$ collected in dental clinics, along with their corresponding orthodontic treatment plans provided by dentists. In this way, we can also obtain corresponding 1257 post-orthodontic intra-oral scanning models $\hat{S}$ accordingly. 
Note that for each of the 138 render-used facial photographs, we randomly select 10 from the pool of 1257 intra-oral scanning models to perform the Render process as mentioned in \ref{subsubsec:render}. Thus we have 1380 pre- and post-orthodontic tooth contours $C_r,\hat{C_r}$, respectively, for training $Align\text{-}Mod$.

\vspace{0.2em}
\noindent {\bf Implementation Details.} The proposed method is implemented in PyTorch with 2 NVIDIA A100 80GB GPU. By iteratively tuning and training $Segm\text{-}Mod$, we utimately choose to assign 0.8 to $w_{dice}$ and 0.2 to $w_{ce}$, along with a weight of 20 for the foreground and a weight of 1 for the background in $\mathcal{L}_{ce}$. In terms of $Align\text{-}Mod$ and $Gen\text{-}Mod$, we set the batch-size of our diffusion model to 60. The learning rate is $5e$-$5$ and we use Exponential Moving Average\citep{haynes2012exponential} with $\beta=0.9999$ to update parameters of diffusion model. Lastly regarding the parameters in Render, we set focal length in camera intrinsic parameters to 213.33, and we use SGD\citep{robbins1951stochastic} optimizer with an initial learning rate of 0.01 and a learning rate scheduler, which reduces the learning rate by a factor of 0.9 every 500 steps.

\subsection{Results}
\begin{figure}[htbp]
  \vspace{-5pt}
 \centering
 \setlength{\abovecaptionskip}{0pt}
 \setlength{\belowcaptionskip}{-13pt}
 \begin{minipage}{0.1614\textwidth}
   \centering
   \includegraphics[width=\textwidth]{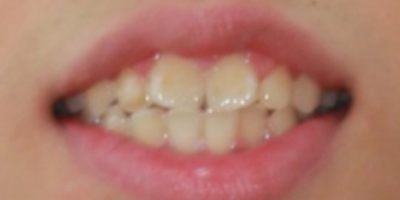}\hfill
   \includegraphics[width=\textwidth]{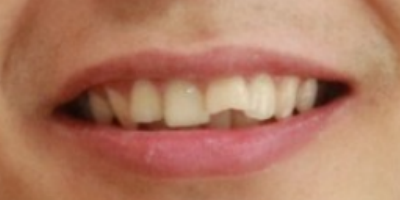}\hfill
   \includegraphics[width=\textwidth]{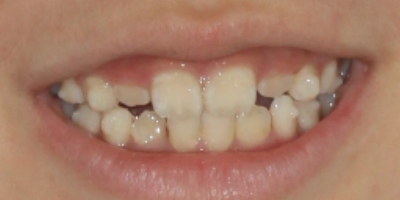}\hfill
   \vspace{-5pt}
   \centerline{\fontsize{5.4}{5}\selectfont (a) mouth region}
 \end{minipage}
 \begin{minipage}{0.1614\textwidth}
   \vspace{-0.25pt}
   \centering
   \includegraphics[width=\textwidth]{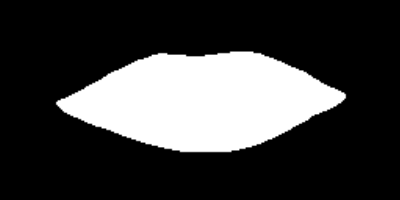}\hfill
   \includegraphics[width=\textwidth]{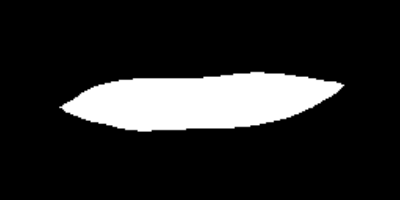}\hfill
   \includegraphics[width=\textwidth]{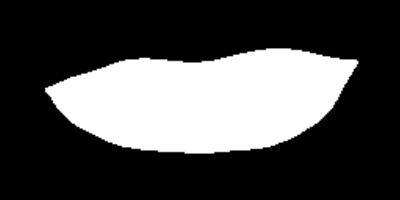}\hfill
   \vspace{-5pt}
   \centerline{\fontsize{5.4}{5}\selectfont (b) oral mask}
 \end{minipage}
 \begin{minipage}{0.1614\textwidth}
   \centering
   \includegraphics[width=\textwidth]{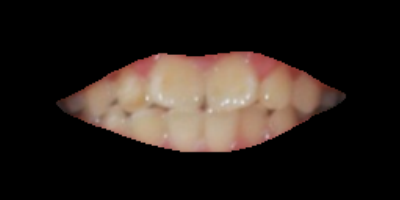}\hfill
   \includegraphics[width=\textwidth]{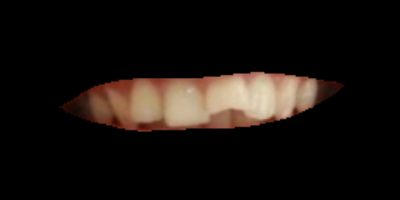}\hfill
   \includegraphics[width=\textwidth]{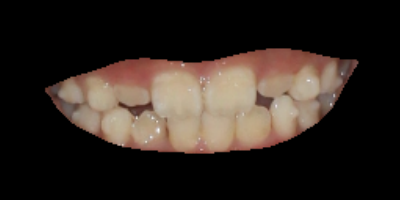}\hfill
   \vspace{-5pt}
   \centerline{\fontsize{5.4}{5}\selectfont (c) oral region}
 \end{minipage}
 \begin{minipage}{0.1614\textwidth}
   \centering
   \includegraphics[width=\textwidth]{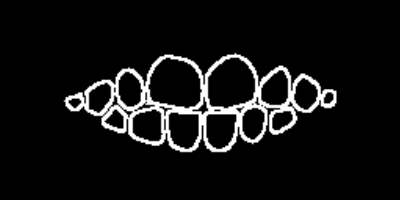}\hfill
   \includegraphics[width=\textwidth]{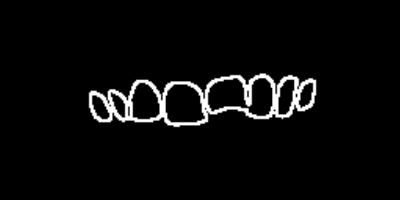}\hfill
   \includegraphics[width=\textwidth]{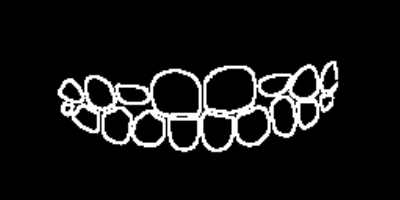}\hfill
   \vspace{-5pt}
   \centerline{\fontsize{5.4}{5}\selectfont (d) segmented tooth contours}
 \end{minipage}
 \begin{minipage}{0.1614\textwidth}
   \centering
   \includegraphics[width=\textwidth]{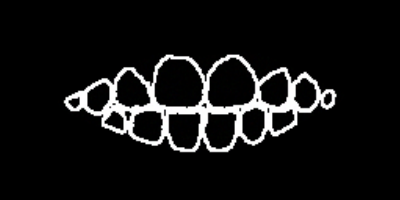}\hfill
   \includegraphics[width=\textwidth]{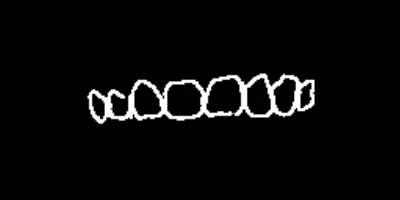}\hfill
   \includegraphics[width=\textwidth]{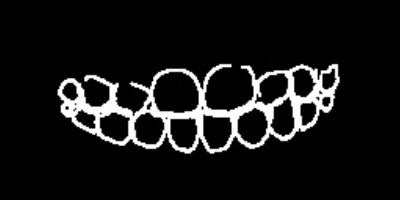}\hfill
   \vspace{-5pt}
   \centerline{\fontsize{5.4}{5}\selectfont (e) aligned tooth contours}
 \end{minipage}
 \begin{minipage}{0.1614\textwidth}
   \centering
   \includegraphics[width=\textwidth]{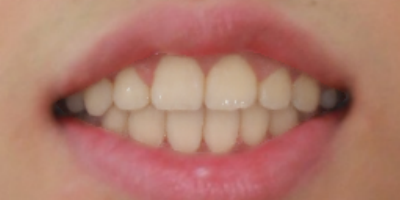}\hfill
   \includegraphics[width=\textwidth]{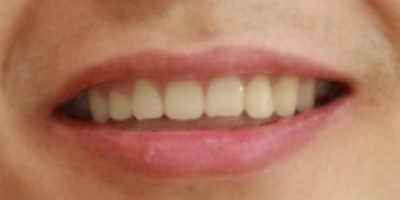}\hfill
   \includegraphics[width=\textwidth]{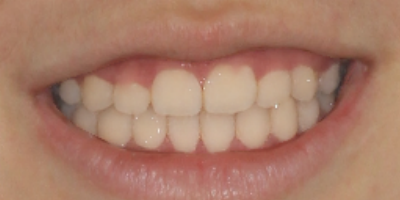}\hfill
   \vspace{-5pt}
   \centerline{\fontsize{5.4}{5}\selectfont (f) predicted mouth region}
 \end{minipage}
 \caption{Inference process. For each detected mouth region $R_i$ (a), we segment to obtain the oral mask $M_i$ (b) and oral region (c). We further obtain tooth contours $C_i$ (d) from our $Segm\text{-}Mod$ and input it into our $Align\text{-}Mod$ to yield well-aligned tooth contours $\hat{C_i}$ (e). We finally predict a mouth region with well-aligned teeth $\hat{R_i}$ (f) through our $Gen\text{-}Mod$.}
 \label{fig:result_process}
\end{figure}

Based on the pre-trained $Align\text{-}Mod$, our three-stage network can infer a facial photograph with well-aligned and aesthetically pleasing teeth $\hat{I_i}$ according to patient's previous photograph $I_i$, without any 3D intra-oral scanning model as input, while still benefiting from the guidance of clinical orthodontic knowledge within 3D structure of intra-oral scanning models. To demonstrate the outstanding results of our method and provide a more detailed view of the inference process and confidence level, we present some testing cases in Figure \ref{fig:result_vis} and Figure \ref{fig:result_process}.

Based on the visual results and inference process we presented, it's evident that our $Segm\text{-}Mod$ has excellent segmentation ability, even with misaligned teeth. Our $Align\text{-}Mod$ also performs great reliability on predicting well-aligned tooth contours, closely based on the pre-trained transformation $\mathcal{T}(\cdot)$ in the image space, which is crucial for clinical orthodontic treatment. Besides, our $Gen\text{-}Mod$ can infer reasonable realistic teeth with similar color and lighting compared with patient's previous teeth photograph and its shooting environment.


\subsection{Comparison}
\begin{figure}[t]
  \setlength{\abovecaptionskip}{0.5pt}
  \setlength{\belowcaptionskip}{-5pt}
  \begin{minipage}{0.5\textwidth}
    \begin{minipage}{0.24\textwidth}
    	  \vspace{1.45pt}
	  \includegraphics[width=\textwidth]{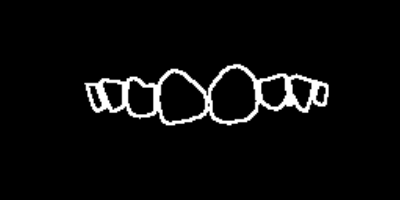}
	  \vspace{3pt}
	  \includegraphics[width=\textwidth]{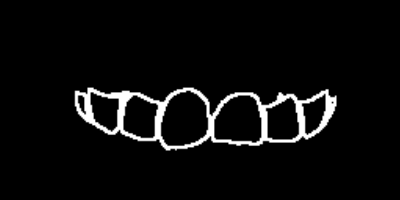}
	  \includegraphics[width=\textwidth]{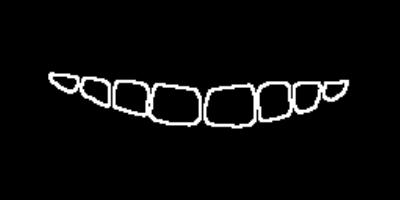}
	  \includegraphics[width=\textwidth]{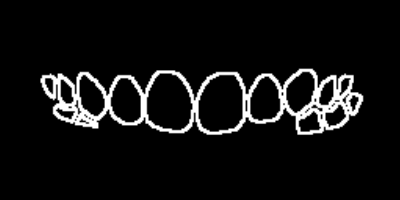}
	  \centerline{\scriptsize Input}
    \end{minipage}
    \begin{minipage}{0.24\textwidth}
    	  \includegraphics[width=\textwidth]{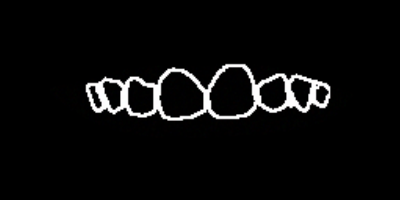}
    	  \vspace{3pt}
	  \includegraphics[width=\textwidth]{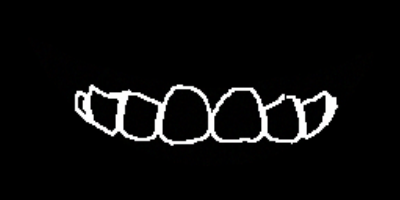}
	  \includegraphics[width=\textwidth]{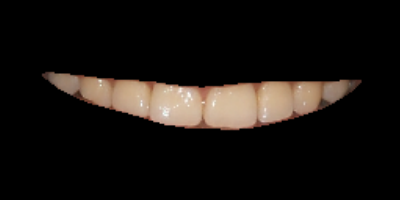}
	  \includegraphics[width=\textwidth]{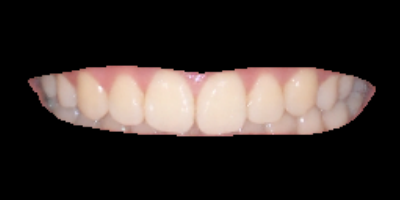}
	  \centerline{\scriptsize Ours}
    \end{minipage}
    \begin{minipage}{0.24\textwidth}
      \vspace{1.45pt}
    	  \includegraphics[width=\textwidth]{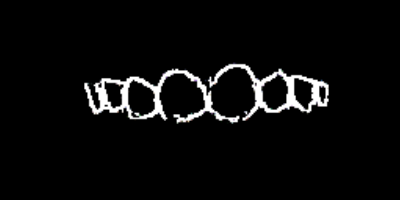}
    	  \vspace{3pt}
	  \includegraphics[width=\textwidth]{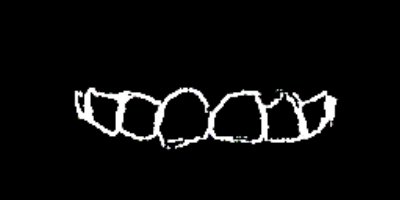}
	  \includegraphics[width=\textwidth]{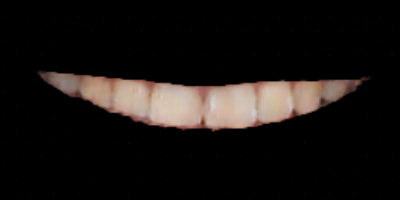}
	  \includegraphics[width=\textwidth]{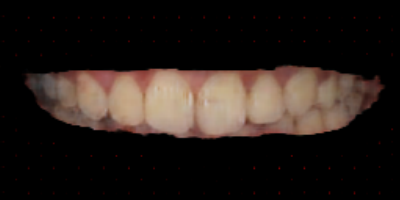}
	  \centerline{\scriptsize Pix2pix GAN}
    \end{minipage}
    \begin{minipage}{0.24\textwidth}
    	  \includegraphics[width=\textwidth]{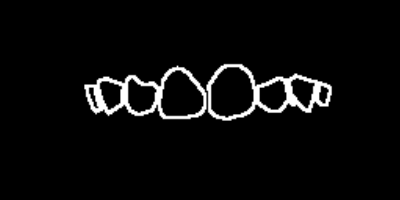}
    	  \vspace{3pt}
	  \includegraphics[width=\textwidth]{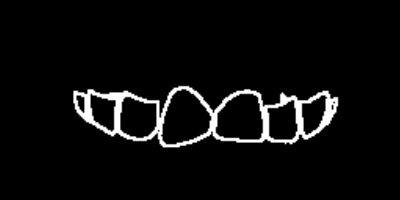}
	  \includegraphics[width=\textwidth]{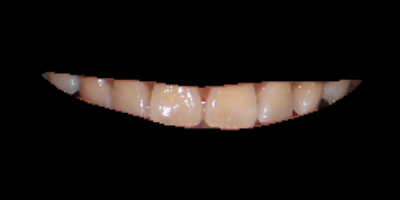}
	  \includegraphics[width=\textwidth]{images/comparison/Out_e44e5fb6d0154666be5363e85209341b}
	  \centerline{\scriptsize GT}
    \end{minipage}
    \captionof{figure}{Qualitative comparisons. The upper two rows are two testing cases of $Align\text{-}Mod$, and the lower two rows are two testing cases of $Gen\text{-}Mod$.}
    \label{fig:comparison}
  \end{minipage}
  \begin{minipage}{0.02\textwidth}\hfill \end{minipage}
  \begin{minipage}{0.45\textwidth}
    \vspace{-12pt}
    \centering
    \begin{tabular}{c|ccc}
      \hline
       & \footnotesize $\mathcal{L}1$ & \footnotesize $\mathcal{L}2$ & \footnotesize LPIPS \\
      \hline
      \multirow{2}{*}{\shortstack[l]{\footnotesize Diffusion Model \\ \footnotesize -based $Align\text{-}Mod$}} & \multirow{2}{*}{\bf \footnotesize 0.053} & \multirow{2}{*}{\bf \footnotesize 0.319} & \multirow{2}{*}{\bf \footnotesize 0.057} \\
      &  &  &\\
      \multirow{2}{*}{\shortstack[l]{\footnotesize Pix2pix GAN \\ \footnotesize -based $Align\text{-}Mod$}} & \multirow{2}{*}{\footnotesize 0.061} & \multirow{2}{*}{\footnotesize 0.343} & \multirow{2}{*}{\footnotesize 0.104} \\
      &  &  &\\
      \hline
      \multirow{2}{*}{\shortstack[l]{\footnotesize Diffusion Model \\ \footnotesize -based $Gen\text{-}Mod$}} & \multirow{2}{*}{\bf \footnotesize 0.029} & \multirow{2}{*}{\bf \footnotesize 0.093} & \multirow{2}{*}{\bf \footnotesize 0.038} \\
      &  &  &\\
      \multirow{2}{*}{\shortstack[l]{\footnotesize Pix2pix GAN \\ \footnotesize -based $Gen\text{-}Mod$}} & \multirow{2}{*}{\footnotesize 0.047} & \multirow{2}{*}{\footnotesize 0.148} & \multirow{2}{*}{\footnotesize 0.107} \\
      &  &  &\\
      \hline
    \end{tabular}
    \captionof{table}{Quantitative comparisons between different methods on Testing Dataset.}
    \label{tab:comparison}
  \end{minipage}
\end{figure}

Additionally, we qualitatively compare our tooth alignment network with Pix2pix GAN\citep{isola2017image,goodfellow2020generative}, especially for $Align\text{-}Mod$ and $Gen\text{-}Mod$, and the comparison results are visually shown in Figure \ref{fig:comparison}. It is shown that our Diffusion Model-based methods are more capable than Pix2pix GAN-based methods, with more reasonable alignment prediction and more realistic tooth color and lighting. As shown in Table \ref{tab:comparison}, we quantitatively evaluate our proposed $Align\text{-}Mod$ and $Gen\text{-}Mod$ to make comparisons with pix2pix GAN using pixel-wise $\mathcal{L}1$, $\mathcal{L}2$ and LPIPS error\citep{johnson2016perceptual,zhang2018unreasonable}. $\mathcal{L}1$ and $\mathcal{L}2$ are commonly used pixel-wise metrics for quantifying discrepancies between generated results and the target, whereas LPIPS is a perceptual metric which calculates the perceptual distance and visual similarity between images. It is shown that our method are consistently better than Pix2pix GAN-based methods considering pixel-wise metrics.

\subsection{Ablation Study}
\begin{figure}[t]
  \setlength{\abovecaptionskip}{0.5pt}
  \setlength{\belowcaptionskip}{-10pt}
  \begin{minipage}{0.38\textwidth}
    \centering
    \begin{tabular}{l|c|c}
      & \multirow{2}{*}{\shortstack[l]{\footnotesize facial \\ \footnotesize skin color}} &  \multirow{2}{*}{\shortstack[l]{\footnotesize intra-oral \\ \footnotesize highlights}}\\
      & & \\
      \hline
      \footnotesize Ablation \uppercase\expandafter{\romannumeral 1} & & \\
      \footnotesize Ablation \uppercase\expandafter{\romannumeral 2} & \checkmark & \\
      \footnotesize Ablation \uppercase\expandafter{\romannumeral 3} & & \checkmark \\
      \footnotesize Ablation \uppercase\expandafter{\romannumeral 4} & \checkmark & \checkmark \\
    \end{tabular}
    \captionof{table}{Four groups of ablation experiments.}
    \label{tab:ablation}
  \end{minipage}
  \begin{minipage}{0.02\textwidth}\end{minipage}
  \begin{minipage}{0.62\textwidth}
    \centering
    \begin{minipage}{0.19\textwidth}
	    \includegraphics[width=\textwidth]{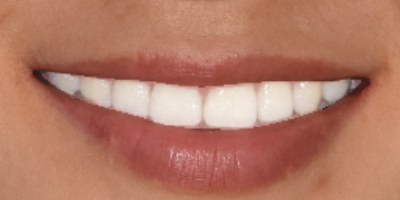}
	    \includegraphics[width=\textwidth]{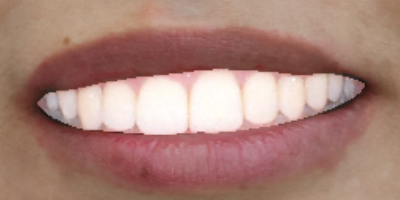}  
	    \includegraphics[width=\textwidth]{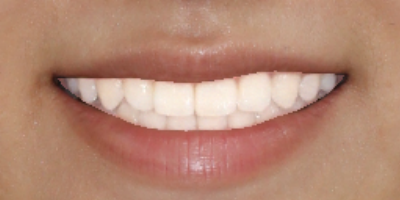} 
	    	\centerline{\scriptsize Ablation \uppercase\expandafter{\romannumeral 1}}
    \end{minipage}
    \hspace{-4pt}
    \begin{minipage}{0.19\linewidth}
 	    \includegraphics[width=\textwidth]{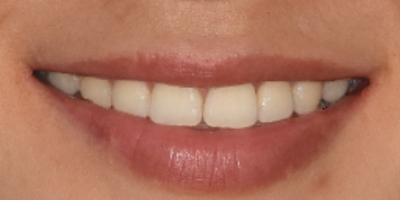}
	    \includegraphics[width=\textwidth]{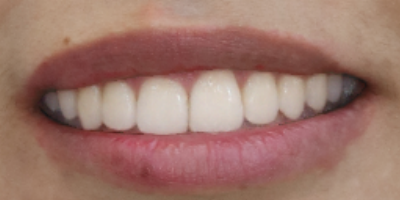} 
	    \includegraphics[width=\textwidth]{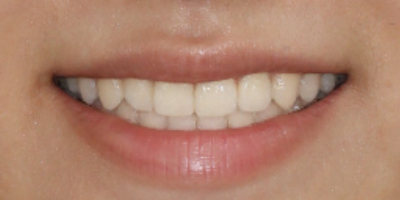}
	 	\centerline{\scriptsize Ablation \uppercase\expandafter{\romannumeral 2}}
    \end{minipage}
    \hspace{-4pt}
    \begin{minipage}{0.19\linewidth}
	    \includegraphics[width=\textwidth]{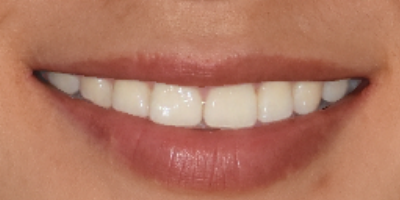}
	    \includegraphics[width=\textwidth]{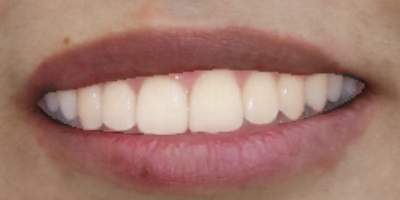}	
	    \includegraphics[width=\textwidth]{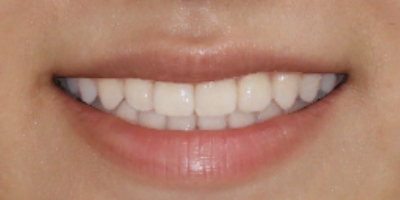}
	    \centerline{\scriptsize Ablation \uppercase\expandafter{\romannumeral 3}}  
    \end{minipage}
    \hspace{-4pt}
    \begin{minipage}{0.19\linewidth}
 	    \includegraphics[width=\textwidth]{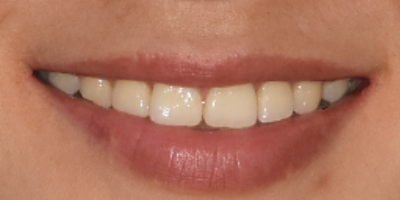}
	    \includegraphics[width=\textwidth]{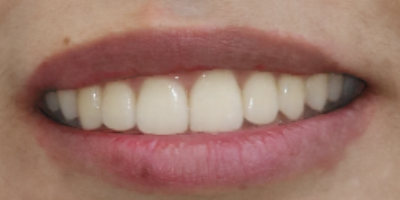}	
	    \includegraphics[width=\textwidth]{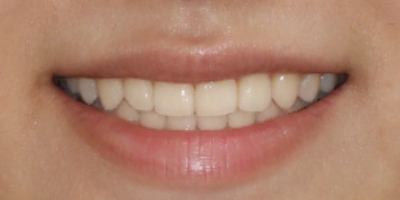}
	    	\centerline{\scriptsize Ablation \uppercase\expandafter{\romannumeral 4}}
    \end{minipage}
    \hspace{-4pt}
    \begin{minipage}{0.19\linewidth}
 	    \includegraphics[width=\textwidth]{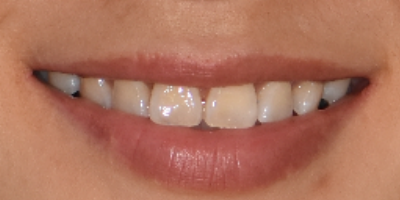}
	    \includegraphics[width=\textwidth]{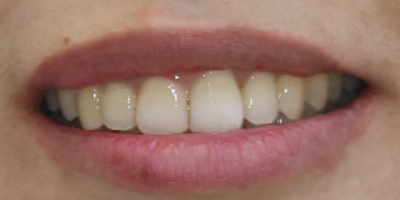}
	    \includegraphics[width=\textwidth]{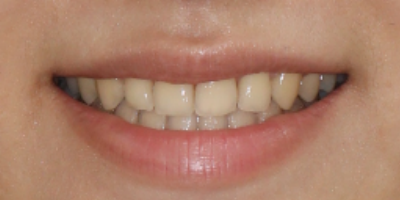}
	    	\centerline{\scriptsize GT}	  
    \end{minipage}
    \captionof{figure}{Visual results of four ablation experiments.}
    \label{fig:ablation}
  \end{minipage}
\end{figure}

We have mentioned in Subsection \ref{subsec:generation} that, in order to make our $Gen\text{-}Mod$ yield more realistic tooth color and environmental lighting, we introduce intra-oral highlights $L_i$ and facial skin color $K_i$ and concatenate them together with well-aligned tooth contours $\hat{C_i}$ and intra-oral Gaussian noise $G_i$ as guidance. To validate the effectiveness of these two condition images guiding $Gen\text{-}Mod$, we conduct an ablation study to demonstrate their effectiveness. We design four groups of ablation experiments as shown in Table \ref{tab:ablation}, all using Diffusion Model-based generative network and the same datasets for training. Visual results of four ablation experiments are shown in Figure \ref{fig:ablation}.

Specifically, Ablation \uppercase\expandafter{\romannumeral 1} includes neither of the two condition images, resulting in the poorest generation performance. Ablation \uppercase\expandafter{\romannumeral 2} includes facial skin color, resulting in better tooth color but lighting information is lost compared to the original image. Ablation \uppercase\expandafter{\romannumeral 3} adds intra-oral highlights, significantly restoring the environmental lighting but less realistic tooth color. Ablation \uppercase\expandafter{\romannumeral 4} has intuitively the best generation performance, with both realistic facial skin color and intra-oral highlights.

\subsection{User Study}
To further demonstrate the reliability and credibility of our method, we conduct a user study that we invite 30 individuals to vote for assessing the alignment and authenticity of photographs (only concentrated on mouth region) generated by our method. Specifically, for assessing alignment, we randomly select 10 generated facial photographs and 10 photographs from patients who have received orthodontic treatments. Participants are asked to rate the alignment of teeth in the photographs on a scale of 1 to 5, with higher scores indicating better alignment. Similarly, for assessing authenticity, we randomly select 10 generated facial photographs and 10 real ones, and ask participants to vote on whether they are real or fake. Table \ref{tab:user} has shown the average alignment scores and average percentage of photographs being classified to "real". Compared with well-aligned or real photographs, photographs generated by our method achieve high scores in terms of both alignment and authenticity, only slightly lower than scores of the well-aligned teeth on real photographs. 

\begin{table}[t]
  \centering
  \setlength{\abovecaptionskip}{0.5pt}%
  \setlength{\belowcaptionskip}{-10pt}%
  \begin{tabular}{c|c|c|c}
    & \makebox[0.2\textwidth][c]{Well-aligned photos} & \makebox[0.2\textwidth][c]{Real photos} & \makebox[0.2\textwidth][c]{Ours} \\
    \hline
    Alignment & 3.84 & \diagbox{} & 3.82 \\
    \hline
    Authenticity & \diagbox{} & 72.67\% & 65.00\% \\
    \hline
  \end{tabular}
  \caption{Voting results of user study. First row represents the average score regarding alignment, second row represents the average percentage of photographs being classified as "real".}
  \label{tab:user}
\end{table}

\section{Discussion} \label{sec:discussion}
In this work, we propose a 3D structure-guided tooth alignment network to effectively generate orthodontic comparison photographs. According to the experimental results above, our method utilizes 3D dental models to learn the orthodontic knowledge in the image space. The 3D structure successfully guides the learning and prediction of our network, giving our method practical clinical significance. Additionally, we introduce Diffusion Model into the task of orthodontic comparison photograph generation and have shown the great power of Diffusion Model in our 
$Align\text{-}Mod$ and $Gen\text{-}Mod$.
Importantly, different from state-of-the-art methods\citep{lingchen2020iorthopredictor,chen2022orthoaligner} in the field, our method can still incorporate clinical orthodontic knowledge into the network without requiring additional input of dental models. This demonstrates that our method is much more clinically practical, user-friendly and applicable within orthodontic industry.

Our method, however, is not without limitations. For example, our network cannot handle with several cases, such as that teeth are too misaligned and patients smile too widely. Moreover, our method cannot take collision and occlusal relationship into consideration since our method is just performed in the image space. In future, we plan to first reconstruct 3D tooth models from 2D photograph and then make the tooth alignment.

\section{Conclusion} \label{sec:conclusion}
In this paper, we have designed a 3D structure-guided network to infer a facial photograph with well-aligned and aesthetically pleasing teeth based on the patient's previous facial photograph. Our method stands out from existing approaches as it can learn the clinical orthodontic knowledge based on 3D intra-oral scanning models, making our method highly reliable and potentially applicable in clinical practice.

\section*{Acknowledgements}
This work was supported in part by NSFC grants (No. 6230012077). 

\bibliography{egbib}
\end{document}